\documentclass[conference]{IEEEtran}
\IEEEoverridecommandlockouts
\usepackage[noadjust]{cite}
\usepackage{amsmath,amssymb,amsfonts}
\usepackage{algorithm}
\usepackage{algpseudocode}
\usepackage{graphicx}
\usepackage{textcomp}
\usepackage{xcolor}
\usepackage{caption}
\usepackage{lipsum}
\usepackage{subcaption}
\def\BibTeX{{\rm B\kern-.05em{\sc i\kern-.025em b}\kern-.08em
    T\kern-.1667em\lower.7ex\hbox{E}\kern-.125emX}}
    
\begin{document}

% Add a command to place the copyright notice at the top
\makeatletter
\def\@maketitle{%
  % First add the copyright notice
  \noindent{\footnotesize \copyright~2025 IEEE. Personal use of this material is permitted. Permission from IEEE must be obtained for all other uses, in any current or future media, including reprinting/republishing this material for advertising or promotional purposes, creating new collective works, for resale or redistribution to servers or lists, or reuse of any copyrighted component of this work in other works.}
  
  \noindent{\footnotesize Accepted for publication at 2025 IEEE International Conference on Communications (ICC), Montreal, Canada, June 8-12, 2025.}
  
  \vspace{0.5cm}
  
  % Then place the title content with reduced spacing
  \begingroup
  \centering
  \vskip0.5em%
  {\Huge\@title\par}%
  \vskip1.0em%
  \@author
  \par
  \endgroup
  \vskip1em
}
\makeatother

\title{AdaFortiTran: An Adaptive Transformer Model for Robust OFDM Channel Estimation}

\author{Berkay Guler and Hamid Jafarkhani\thanks{This work was supported
in part by the NSF Award CNS-2229467.}\\
        \textit{Center for Pervasive Communications and Computing}\\
        \textit{University of California, Irvine}
}

% \author{
%     \IEEEauthorblockN{Berkay Guler and Hamid Jafarkhani \thanks{This work was supported
% in part by the NSF Award CNS-2229467. }}
%     \IEEEauthorblockA{
%         \textit{Center for Pervasive Communications and Computing} \\
%         \textit{University of California, Irvine}\\
        %CA, USA \\
%        gulerb@uci.edu}
% }
%\and
%    \IEEEauthorblockN{Hamid Jafarkhani}
%    \IEEEauthorblockA{
%        \textit{Electrical Engineering and Computer Science Department} \\
%        \textit{University of California, Irvine}\\
%        CA, USA \\
%        hamidj@uci.edu}
% }

\maketitle

\begin{abstract}

Deep learning models for channel estimation in Orthogonal Frequency Division Multiplexing (OFDM) systems often suffer from performance degradation under fast-fading channels and low-SNR scenarios. To address these limitations, we introduce the Adaptive Fortified Transformer (AdaFortiTran), a novel model specifically designed to enhance channel estimation in challenging environments. Our approach employs convolutional layers that exploit locality bias to capture strong correlations between neighboring channel elements, combined with a transformer encoder that applies the global Attention mechanism to channel patches. This approach effectively models both long-range dependencies and spectro-temporal interactions within single OFDM frames. We further augment the model's adaptability by integrating nonlinear representations of available channel statistics SNR, delay spread, and Doppler shift as priors. A residual connection is employed to merge global features from the transformer with local features from early convolutional processing, followed by final convolutional layers to refine the hierarchical channel representation. Despite its compact architecture, AdaFortiTran achieves up to 6 dB reduction in mean squared error (MSE) compared to state-of-the-art models. Tested across a wide range of Doppler shifts (200-1000 Hz), SNRs (0 to 25 dB), and delay spreads (50-300 ns), it demonstrates superior robustness in high-mobility environments.
\end{abstract}

\begin{IEEEkeywords}
channel estimation, OFDM, Transformer, Attention, Deep learning
\end{IEEEkeywords}
\section{Introduction}

Orthogonal Frequency Division Multiplexing (OFDM) is widely used for its resilience to multipath fading and high spectral efficiency. Accurate channel estimation (CE) is vital for optimal OFDM performance, particularly in dynamic wireless environments \cite{ofdm_ce1}. High noise degrades OFDM signal quality, while Doppler shifts and frequency offsets in high-mobility scenarios disrupt subcarrier orthogonality. Delay spread from multipath propagation introduces inter-symbol interference, further affecting signal integrity. These challenges limit the effectiveness of conventional channel estimators, complicating equalization and data recovery. While increasing pilot density helps address these issues, it reduces throughput and spectral efficiency, making advanced adaptive methods essential for handling noise and high-mobility environments \cite{ofdm_ce2, ofdm_ce3}.

Pilot-assisted channel estimation (PA-CE) is the predominant CE method for OFDM and relies on known symbols placed within the transmitted frame. Two common PA-CE methods are the Least Squares (LS) \cite{ofdm_ls} and Linear Minimum Mean Square Error (LMMSE) estimators \cite{ofdm_lmmse}. The LS estimator offers simplicity but does not consider channel statistics, while LMMSE provides better adaptability using second-order statistics at the cost of higher complexity.

Deep learning-based channel estimation (DL-CE) has emerged as a promising alternative to traditional methods, gaining popularity for its superior performance \cite{deep_ce1, deep_ce2}. An early work proposed a two-step method where a convolutional neural network (CNN) upsamples the LS channel estimates at pilot positions, followed by a denoising CNN to refine the upsampled estimates \cite{channelnet}. A later work introduced residual blocks to connect CNN-generated feature maps, improving performance and efficiency \cite{reesnet}. Another approach employed a two-stage network comprising one CNN operating in the spatial-frequency domain and another in the angle-delay domain \cite{dualcnn}.

Following the success of CNNs, recurrent neural networks (RNNs) and their variants have emerged as powerful tools for CE. In \cite{cnn_lstm}, Long Short-Term Memory (LSTM) networks were employed to capture temporal features, while a CNN extracted spatial features. Another study utilized Gated Recurrent Units (GRUs) to refine LS estimates \cite{gru}. More recently, Bidirectional GRUs in conjunction with convolutional layers have been applied to process channel elements along the frequency axis \cite{bigru}.

Following the widespread adoption of Transformers \cite{tf} in language \cite{bert} and vision tasks \cite{vit}, several transformer-based CE methods have been developed. The first application of transformers for CE relied on the Transformer encoder from the original paper \cite{tf} and used features from Discrete Fourier Transform estimates at pilot positions, extracted by a 1D-CNN network, as the input \cite{tf1}. A hybrid architecture combined a transformer-based encoder with a residual CNN \cite{tf2}. Recently, a relatively new transformer variant, Vision Transformer \cite{vit}, together with channel tokens outperformed previous DL-CE methods \cite{tf3}.

In this paper, we present Adaptive Fortified Transformer (AdaFortiTran), which exploits strong correlations between neighboring channel elements through the locality bias of CNNs. We utilize the translation invariance of 2D convolutional filters to capture features appearing across the channel. AdaFortiTran processes sequences of tiny channel patches with a transformer-based encoder to extract deep spectro-temporal features. Unlike Vision Transformer \cite{vit}, which employs large patches like \((32\times32)\), we use \((3\times2)\) patches to achieve higher dual-domain resolution with finer-grained attention. Smaller patches ensure better granularity in capturing variations across time and frequency, enhancing estimation performance in high-mobility environments. Our model combines shallow local features from the initial convolutional processing with deep features from the transformer layers. Finally, the channel restoration module reconstructs the channel by refining hierarchical features.

Inspired by \cite{tf3}, we incorporate channel state encodings to enhance the adaptivity of our model by conditioning attention on channel statistics. We explore the trade-off between model size and performance by varying the number of transformer layers and present our findings. We verify AdaFortiTran's performance across various scenarios using channels from 3rd Generation Partnership Project (3GPP) specification \cite{3gpp_tdl}. AdaFortiTran outperforms leading DL-CE models Ce-ViT \cite{tf3} and SisRafNet \cite{bigru} with up to 6~dB MSE improvement while maintaining smaller model sizes, as presented later in Section \ref{sec:sim}. \footnote{We make AdaFortiTran along with our dataset publicly available at: https://github.com/BerkIGuler/AdaFortiTran.}

\section{System Model and Problem Formulation}
\subsection{System Model}
We consider a single-input single-output (SISO) OFDM system in line with the current 5G NR specification \cite{3gpp_nr}. We use 15~kHz subcarrier spacing, 10 resource blocks, and 12 subcarriers per block, resulting in \(N_f = 120\) subcarriers and \(N_t = 14\) OFDM symbols per frame, with a total bandwidth of 1.8~MHz. QPSK-modulated signals are transmitted over a TDL-A channel \cite{3gpp_tdl} and sampled at a rate of 3.84~MHz. After cyclic prefix removal and Discrete Fourier Transform, the received symbol $y_{n,k} \in \mathbb{C}$ at subcarrier \(n\) and time index \(k\) is given by:
\begin{equation}
y_{n,k} = h_{n,k} x_{n,k} + \epsilon_{n,k},    
\end{equation}
where \(h_{n,k}\in \mathbb{C}\) and \(x_{n,k} \in \mathbb{C}\) represent the channel gain and transmitted symbol, respectively, and \(\epsilon_{n,k} \sim \mathcal{N}(0, \sigma^2)\) is the Gaussian noise. We adopt a lattice-type pilot positioning scheme where pilot symbols are inserted at the \(3^{\text{rd}}\) and \(12^{\text{th}}\) time indices. In the frequency domain, pilots are placed every $N$ subcarriers. We denote the sets of pilot indices along subcarriers and OFDM symbols with \(N_{f,p}\) and \(N_{t,p}\), respectively.
\subsection{Traditional CE Methods}
\subsubsection{LS Estimator}
Define \(\mathcal{P} = N_{f,p} \times N_{t,p}\) as the set of pairs of pilot indices. The LS estimate is obtained as follows:
\begin{equation}
\mathbf{\hat{h}}_p^{\text{LS}} = \underset{\mathbf{h}_p}{\arg\min} \|\mathbf{y}_p - \mathbf{X}_p \mathbf{h}_p\|_2^2 = \mathbf{X}_p^{-1} \mathbf{y}_p,
\label{eq:ls_estimate}
\end{equation}
where \(\mathbf{\hat{h}}_p^{\text{LS}}\), $\mathbf{h}_p$, \(\mathbf{y}_p \in \mathbb{C}^{|\mathcal{P}|\times1}\) represent the LS channel estimate, channel, and received symbols all at pilot positions, respectively. \(\mathbf{X}_p \in \mathbb{C}^{|\mathcal{P}|\times|\mathcal{P}|}\) is a diagonal matrix, and \(\text{diag}(\mathbf{X}_p) = \mathbf{x}_p \in \mathbb{C}^{|\mathcal{P}|\times1}\) is the transmitted symbols at pilot positions. Various interpolation schemes can be applied to obtain \(\mathbf{\hat{H}^{\text{LS}}} \in \mathbb{C}^{N_f \times N_t}\) from \(\mathbf{\hat{h}}_p^{\text{LS}}\).

\subsubsection{LMMSE Estimator}
When second-order channel statistics and noise estimates are available, the minimum mean square error (MMSE) estimator offers improved performance. However, MMSE computation involves multiple matrix inversions, making it computationally expensive. In contrast, the LMMSE estimator \cite{ofdm_lmmse} requires a single matrix inversion, sacrificing some performance for efficiency. The formula for the LMMSE is given by:
\begin{equation}
\mathbf{\hat{h}}_{\text{LMMSE}} = \mathbf{R}_{h h_p} \left( \mathbf{R}_{h_p h_p} + \sigma^2 \mathbf{I} \right)^{-1} \mathbf{\hat{h}}_p^{\text{LS}},
\label{eq:lmmse_estimate}
\end{equation}
where $\mathbf{\hat{h}}_{\text{LMMSE}}\in \mathbb{C}^{(N_fN_t)\times1}$, $\mathbf{R}_{h h_p}=\mathbb{E}[\mathbf{h}\mathbf{h}_p^H] \in \mathbb{C}^{(N_fN_t)\times|\mathcal{P}|}$, $\mathbf{R}_{h_p h_p}=\mathbb{E}[\mathbf{h}_p\mathbf{h}_p^H]\in\mathbb{C}^{|\mathcal{P}|\times|\mathcal{P}|}$, $\mathbf{I} \in \mathbb{C}^{|\mathcal{P}|\times|\mathcal{P}|}$, and $\mathbf{h}\in \mathbb{C}^{(N_fN_t)\times1}$ are LMMSE estimate of the channel, the cross-correlation matrix between the channel and the channel at pilot positions, the auto-correlation matrix of the channel at pilot positions, the identity matrix, and the vectorized channel, respectively.  Here, $(\cdot)^{H}$ is the Hermitian operator. In this work, we use the sample mean estimator to estimate $\sigma^2$, $\mathbb{E}[\mathbf{h}\mathbf{h}_p^H]$, and $\mathbb{E}[\mathbf{h}_p\mathbf{h}_p^H]$.
\subsection{Problem Formulation}
We adopt a data-driven approach to learn a mapping \(f:\mathbb{C}^{N_{f,p} \times N_{t,p}} \rightarrow \mathbb{C}^{N_f \times N_t}\). The function \(f\) is parameterized by \(\boldsymbol{\theta}\in\mathbb{R}^d\) and optimized by minimizing the MSE over the dataset \(\mathcal{D}=\{(\mathbf{\hat{h}}_{p_i}^{\text{LS}}, \mathbf{h}_i)\}_{i=1}^K\). The training objective is formulated as:
\begin{equation}
\min_{\boldsymbol{\theta}} \left\{ \frac{1}{K} \sum_{i=1}^K \|f(\mathbf{\hat{h}}_{p_i}^{\text{LS}}; \boldsymbol{\theta}) - \mathbf{h}_i\|_2^2 \right\},
\end{equation}
where we model \(f\) with AdaFortiTran by processing imaginary and real parts of the channel separately and combining the results to obtain \(f(\mathbf{\hat{h}}_{p}^{\text{LS}}; \boldsymbol{\theta})\).
\section{Proposed Method}
\subsection{Network Architecture}
Illustrated in Fig.~\ref{fig:arch}, AdaFortiTran is a novel architecture for channel estimation. The following subsections detail each component of our proposed model.

\subsubsection{Upsampler and Feature Enhancer}
\begin{figure*}[t]
  \centering
  \begin{subfigure}{\textwidth}
      \centering
      \includegraphics[width=\textwidth]{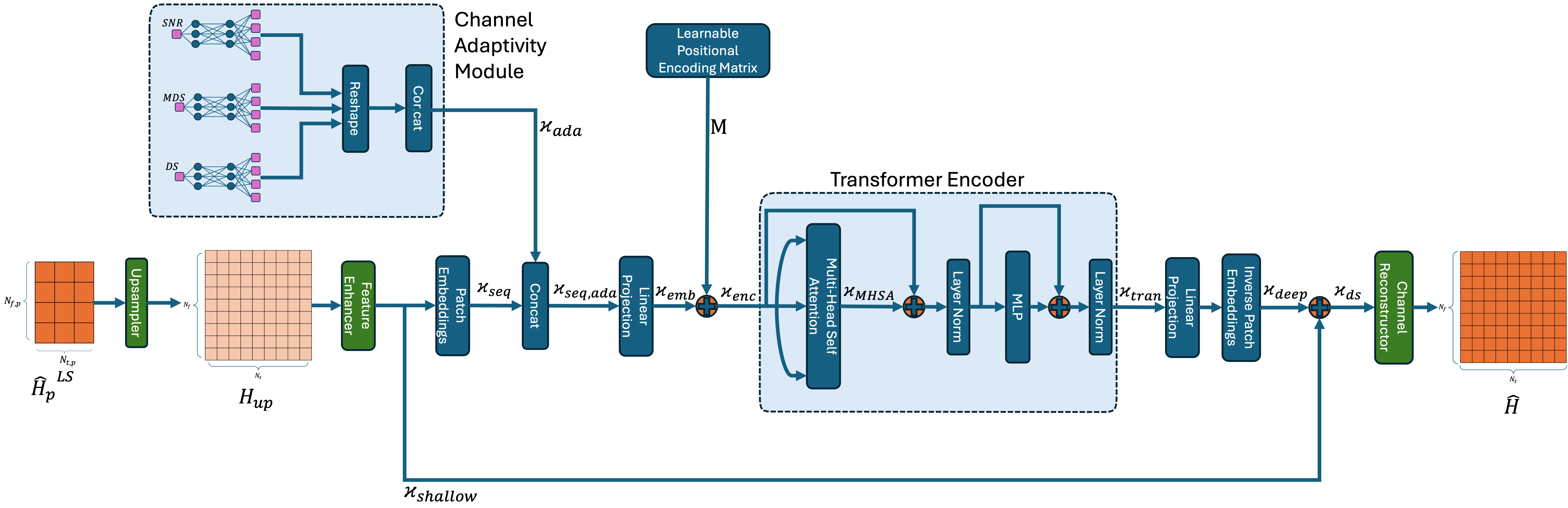}
      \caption{AdaFortiTran Architecture}
      \label{fig:arch}
  \end{subfigure}

  \vspace{0.2cm}

  \begin{subfigure}{0.26\textwidth}
      \centering
      \includegraphics[width=\textwidth]{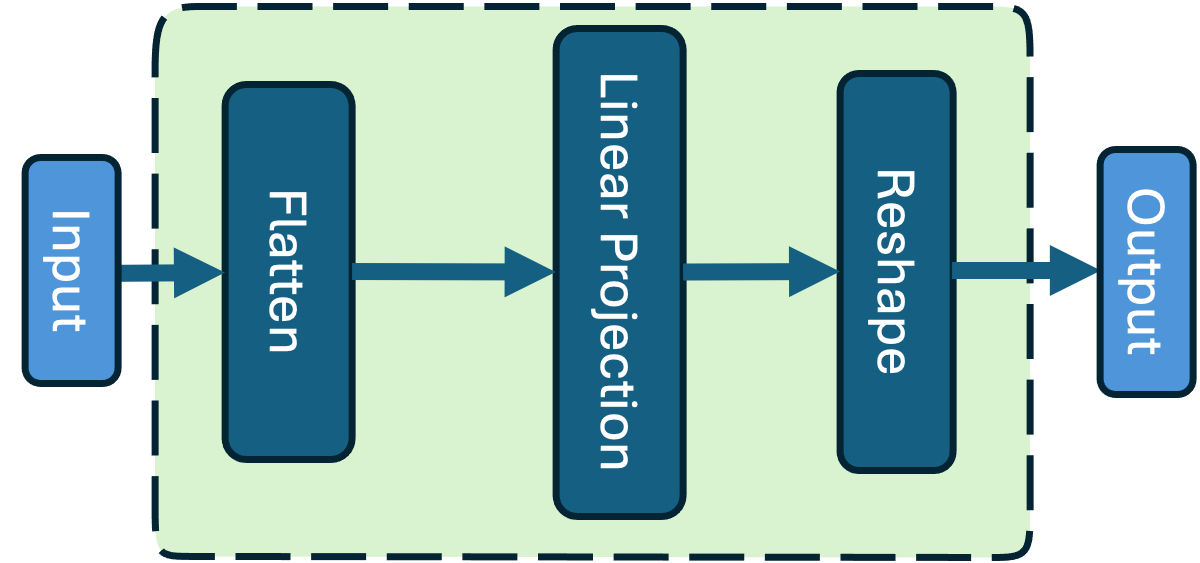}
      \caption{Upsampler Submodule}
      \label{fig:linear_block}
  \end{subfigure}
  \begin{subfigure}{0.68\textwidth}
      \centering
      \includegraphics[width=\textwidth]{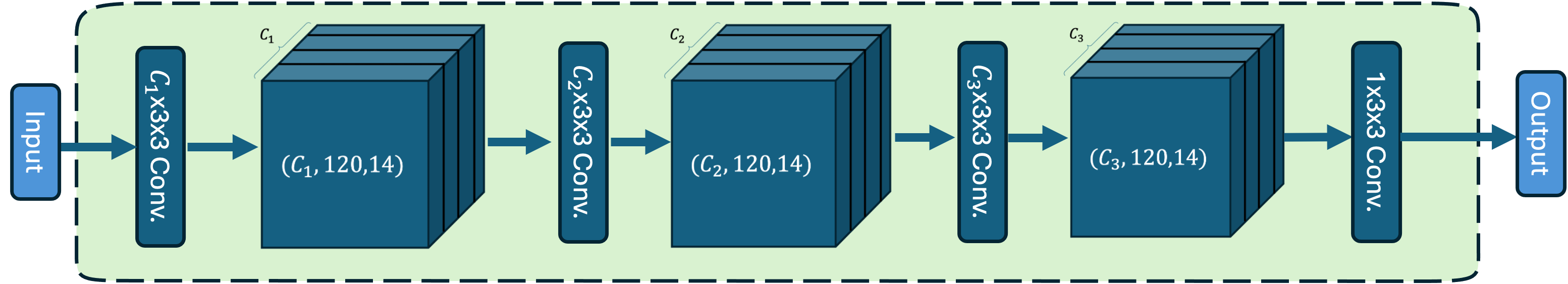}
      \caption{Feature Enhancer and Channel Reconstructor Submodules}
      \label{fig:conv_block}
  \end{subfigure}

  \caption{Architecture and Submodules of AdaFortiTran}
  \label{fig:main_figure}
\end{figure*}

AdaFortiTran takes a channel estimate matrix at pilot positions as input and upsamples it to $\mathbf{H}_{\text{up}} \in \mathbb{R}^{N_f \times N_t}$ through a series of operations. In this work, we specifically use LS estimates $\mathbf{\hat{H}}_p^{\text{LS}} \in \mathbb{R}^{N_{f,p} \times N_{t,p}}$. 
As shown in Fig. \ref{fig:linear_block}, first, $\mathbf{\hat{H}}_p^{\text{LS}}$ is flattened into the vector $\mathbf{\hat{h}}_p^{\text{LS}}$. Then, a linear projection $\mathbf{W}_1 \in \mathbb{R}^{(N_f N_t) \times |\mathcal{P}|}$ is applied along with a bias vector $\mathbf{b}_1 \in \mathbb{R}^{(N_f N_t) \times 1}$, followed by reshaping to obtain $\mathbf{H}_{\text{up}}$. If we ignore the bias term, the linear upsampling performed here is not different from applying a 2D Wiener filter where filter coefficients are learned from data \cite{wiener}. Although another similar work chose to use interpolated LS estimates as input \cite{tf3}, a learned linear upsampling block led to superior performance in our experiments.

The Feature Enhancer Module processes $\mathbf{H}_{\text{up}}$ through a sequence of $C_1=8$, $C_2=32$, and $C_3=8$ convolutional kernels with size $(3 \times 3)$, followed by a single $(3 \times 3)$ convolution, as shown in Fig. \ref{fig:conv_block}. Each convolutional block except the last one is followed by a Rectified Linear Unit (ReLU) activation function to include nonlinearity. We minimized both the number of convolutional kernels and their width to create a shallow feature extractor with a small receptive field to keep the model compact. Therefore, this module produces a shallow feature map $\mathcal{H}_{\text{shallow}} \in \mathbb{R}^{N_f \times N_t}$. The convolutional block aims to inject locality bias and translation equivariance property lacked by transformer. This type of early convolutional processing is shown to help optimize Vision Transformer \cite{cnn_vit}. Another benefit of this block is that its output is used to constitute a hierarchichal representation in the following layers.  

\subsubsection{Channel Adaptivity Module and Patch Embeddings}
The shallow feature map $\mathcal{H}_{\text{shallow}}$ is partitioned into $(3 \times 2)$ patches. These patches are flattened and concatenated to form a sequence of channel vectors $\mathcal{H}_{\text{seq}} \in \mathbb{R}^{(N_f N_t/6) \times 6}$. Choosing a small patch size ensures that correlations among the channel elements are better captured by producing higher resolution attention maps. While this increases the computational complexity of Scaled Dot-product Attention calculation, which has $\mathcal{O}(n^2)$ time complexity, we find this trade-off acceptable. Although more efficient attention variants like \cite{swin} are popular for image processing, we opt for Scaled Dot-product Attention, which is typically used for text processing \cite{bert}. This choice is justified as channel correlations can exist between distant channel elements, the OFDM dimension is fairly manageable compared to natural images, and global patterns in frequency/time domains are crucial for estimation.

The Channel Adaptivity Module (CAM) generates adaptive encodings $\mathcal{H}_{\text{ada}} \in \mathbb{R}^{(N_f N_t/6) \times 6}$ by processing SNR, maximum Doppler shift, and delay spread values through Multi-Layer Perceptrons (MLPs). Each parameter is encoded using separate MLPs with identical structures:
\begin{equation}
\small
\text{MLP}_{i}(x) = \mathbf{W}_{3,i}(\phi(\mathbf{W}_{2,i}(\phi(\mathbf{W}_{1,i} x + \mathbf{b}_{1,i})) + \mathbf{b}_{2,i})) + \mathbf{b}_{3,i},   
\end{equation}
where $\phi$ represents the ReLU nonlinearity, $i \in \{1,2,3\}$ indexes the channel statistic type, $x \in \mathbb{R}$ is the value to encode, $\mathbf{W}_{1,i} \in \mathbb{R}^{7 \times 1}$, $\mathbf{W}_{2,i} \in \mathbb{R}^{42 \times 7}$, $\mathbf{W}_{3,i} \in \mathbb{R}^{(N_f N_t/3) \times 42}$ are projection matrices, and $\mathbf{b}_{1,i} \in \mathbb{R}^{7 \times 1}$, $\mathbf{b}_{2,i} \in \mathbb{R}^{42 \times 1}$, $\mathbf{b}_{3,i} \in \mathbb{R}^{(N_f N_t/3) \times 1}$ are bias terms. It is possible to obtain the input $x$ by probing the channel \cite{tf3}, but we assume that they are already known in this setup. We also include a variant of AdaFortiTran, named FortiTran, without CAM in our analysis, which could be used in the absence of these parameters.
MLP outputs are reorganized to shape $(N_f N_t/6) \times 2$ and concatenated to obtain $\mathcal{H}_{\text{ada}}$. Then, $\mathcal{H}_{\text{seq}}$ is concatenated with $\mathcal{H}_{\text{ada}}$ to form $\mathcal{H}_{\text{seq,ada}} \in \mathbb{R}^{(N_f N_t/6) \times 12}$. After concatenation, half of each flattened channel patch vector contains adaptive elements, forcing the attention calculation to be conditioned on channel statistics.

\subsubsection{Channel Embeddings with Positional Encoding}
Since flattened channel patch vectors do not have a high dimension, we project $\mathcal{H}_{\text{seq,ada}}$ to higher dimension with linear transformation $\mathbf{W}_2 \in \mathbb{R}^{12 \times d_{\text{enc}}}$ and add a bias term $\mathbf{b}_2 \in \mathbb{R}^{d_{\text{enc}} \times 1}$ to obtain $\mathcal{H}_{\text{emb}} \in \mathbb{R}^{(N_f N_t/6) \times d_{\text{enc}}}$. To preserve spatial information regarding the original positions of each patch, following \cite{vit}, we add a learnable positional encoding matrix $\mathbf{M} \in \mathbb{R}^{(N_f N_t/6) \times d_{\text{enc}}}$ to $\mathcal{H}_{\text{emb}}$, element-wise, producing the transformer encoder input $\mathcal{H}_{\text{enc}} \in \mathbb{R}^{(N_f N_t/6) \times d_{\text{enc}}}$. In our experiments, learning the positional encodings provided slightly better performance than using deterministic positional encodings such as a sinusoidal function.

\subsubsection{Transformer Encoder}
\begin{figure}[b]
    \centering
    \includegraphics[width=\columnwidth]{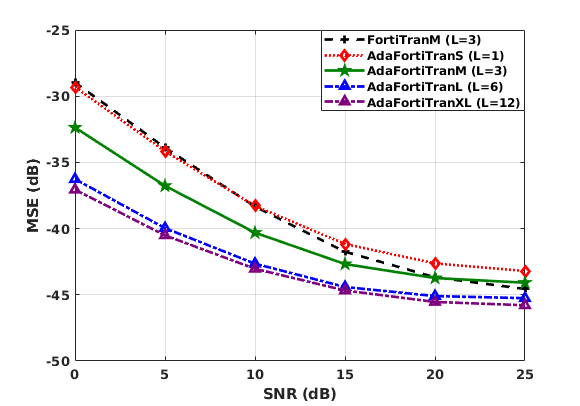}
    \caption{Effect of Channel Adaptivity Module and Number of Transformer Layers $L$}
    \label{fig:layers}
\end{figure}

The transformer-based encoder consists of $L$ layers for deep feature extraction. Within each layer, the Multi-Head Self-Attention (MHSA) block processes $\mathcal{H}_{\text{enc}}$ through $M$ parallel self-attention calculations, producing outputs typically called heads $\{\text{head}_i\}_{i=1}^M$ where each $\text{head}_i \in \mathbb{R}^{(N_f N_t/6) \times (d_{\text{enc}}/M)}$. In our work, different from \cite{tf}, the Scaled Dot-product Attention is computed with bias terms:
\begin{equation}
\mathbf{A}_{\text{ada},i} = \text{softmax}\left(\frac{(\mathcal{H}_{\text{enc}}\mathbf{W}_i^Q + \mathbf{B}_i^Q)
(\mathcal{H}_{\text{enc}}\mathbf{W}_i^K + \mathbf{B}_i^K)^\top}{\sqrt{d_{\text{enc}}/M}}\right),
\end{equation}
\begin{equation}
\text{head}_i = \mathbf{A}_{\text{ada},i}(\mathcal{H}_{\text{enc}}\mathbf{W}_i^V + \mathbf{B}_i^V),
\end{equation}
where $\mathbf{A}_{\text{ada},i} \in \mathbb{R}^{(N_f N_t/6) \times (N_f N_t/6)}$ is the $i$-th adaptive channel attention map and $\mathbf{A}_{\text{ada},i}[k,l]$ represents the relationship strength between the $k$-th and $l$-th channel patches. Here, $\mathbf{W}_i^Q$, $\mathbf{W}_i^K$, $\mathbf{W}_i^V \in \mathbb{R}^{d_{\text{enc}} \times (d_{\text{enc}}/M)}$ and $\mathbf{B}_i^Q$, $\mathbf{B}_i^K$, $\mathbf{B}_i^V \in \mathbb{R}^{(N_f N_t/6) \times (d_{\text{enc}}/M)}$ are projection matrices and bias matrices of the $i$-th head. The heads are concatenated to the shape $(N_f N_t/6) \times d_{\text{enc}}$ and projected with a linear transformation $\mathbf{W}^O \in \mathbb{R}^{d_{\text{enc}} \times d_{\text{enc}}}$ and a bias matrix $\mathbf{B}^O \in \mathbb{R}^{(N_f N_t/6) \times d_{\text{enc}}}$, resulting in the MHSA output $\mathcal{H}_{\text{MHSA}} \in \mathbb{R}^{(N_f N_t/6) \times d_{\text{enc}}}$. The output is processed through additional transformations:
\begin{equation}
  \mathbf{Z} = \text{LN}(\mathcal{H}_{\text{MHSA}} + \mathcal{H}_{\text{enc}}),
\end{equation}
\begin{equation}
  \mathcal{H}_{\text{tran}} = \text{LN}(\mathbf{Z} + \text{MLP}(\mathbf{Z})),
\end{equation}
where $\text{LN}(\cdot)$ represents Layer Normalization \cite{layer_norm} and $\text{MLP}(\mathbf{Z}) = \psi(\mathbf{Z}\mathbf{W}_{t,1} + \mathbf{B}_{t,1})\mathbf{W}_{t,2} + \mathbf{B}_{t,2}$, with $\mathbf{W}_{t,1} \in \mathbb{R}^{d_{\text{enc}} \times 2d_{\text{enc}}}$ and $\mathbf{W}_{t,2} \in \mathbb{R}^{2d_{\text{enc}} \times d_{\text{enc}}}$ serving as linear projections, $\mathbf{B}_{t,1} \in \mathbb{R}^{(N_f N_t/6) \times 2d_{\text{enc}}}$ and $\mathbf{B}_{t,2} \in \mathbb{R}^{(N_f N_t/6) \times d_{\text{enc}}}$ as bias matrices, and $\psi(\cdot)$ denoting the Gaussian Error Linear Unit (GELU). We set $L=6$, $M=4$, and $d_{\text{enc}} = 32$ to balance the performance-complexity trade-off. As shown in Fig. \ref{fig:layers}, for $L > 6$, the marginal performance improvement from additional layers becomes negligible.

\subsubsection{Feature Fusion and Channel Reconstruction}
The transformer output undergoes a linear projection via $\mathbf{W}_3 \in \mathbb{R}^{d_{\text{enc}} \times 6}$ with bias $\mathbf{B}_3 \in \mathbb{R}^{(N_f N_t/6) \times 6}$ to return to shape $(N_f N_t/6) \times 6$, followed by patch partition inversion to produce $\mathcal{H}_{\text{deep}} \in \mathbb{R}^{N_f \times N_t}$. We combine this with the shallow features through element-wise addition, $\mathcal{H}_{\text{ds}} = \mathcal{H}_{\text{deep}} + \mathcal{H}_{\text{shallow}}$, to obtain a hierarchical feature map that contains both low and high frequency details. This integration of shallow and deep features has been widely applied in image super-resolution tasks, which share similar objectives with channel estimation in this context \cite{swinir, drct}. Finally, the Channel Reconstructor Module refines $\mathcal{H}_{\text{ds}}$ through several convolutional layers to produce the final channel estimate $\mathbf{\hat{H}}$. The Channel Reconstructor uses the same shallow architecture as the Feature Enhancer, shown in Fig. \ref{fig:conv_block}, which suffices for local refinement.

\section{Simulation Results}\label{sec:sim}
\begin{table}[b]
\renewcommand{\arraystretch}{1.3}
\caption{Size of different Models}
\label{tab:comparison}
\centering
\begin{tabular}{c||c}
\hline
\bfseries Model & \bfseries Number of Parameters\\
\hline\hline
SisRafNet & 0.43M\\
Ce-ViT & 0.23M\\
AdaFortiTranS (L=1) & 0.18M\\
AdaFortiTranM (L=3) & 0.20M\\
AdaFortiTranL (L=6) & 0.22M\\
AdaFortiTranXL (L=12) & 0.28M\\
FortiTranM (L=3) & 0.12M\\
\hline
\end{tabular}
\end{table}

We compare AdaFortiTran with Ce-ViT \cite{tf3}, SisRafNet \cite{bigru}, LS, LMMSE, and a linear model to evaluate its performance. We denote different versions of AdaFortiTran with suffixes S, M, L, and XL, corresponding to models with $L=1$, $L=3$, $L=6$, and $L=12$ transformer layers, respectively. We also include models without the Channel Adaptivity Module, named FortiTran with the same suffix notation. We extensively test the models' robustness under high-fading channels and low SNR scenarios. We investigate the models' performance across different pilot placements and demonstrate how AdaFortiTran's performance scales with the number of transformer layers.

\begin{figure*}[t]
    \centering
    \begin{subfigure}{0.32\textwidth}
        \centering
        \includegraphics[width=\textwidth]{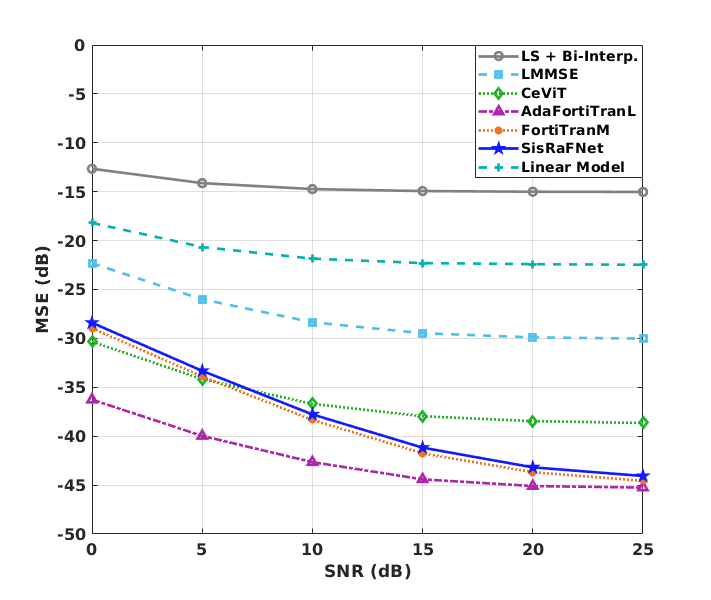}
        \caption{MSE vs. Signal to Noise Ratio}
        \label{fig:snr}
    \end{subfigure}
    \hfill
    \begin{subfigure}{0.32\textwidth}
        \centering
        \includegraphics[width=\textwidth]{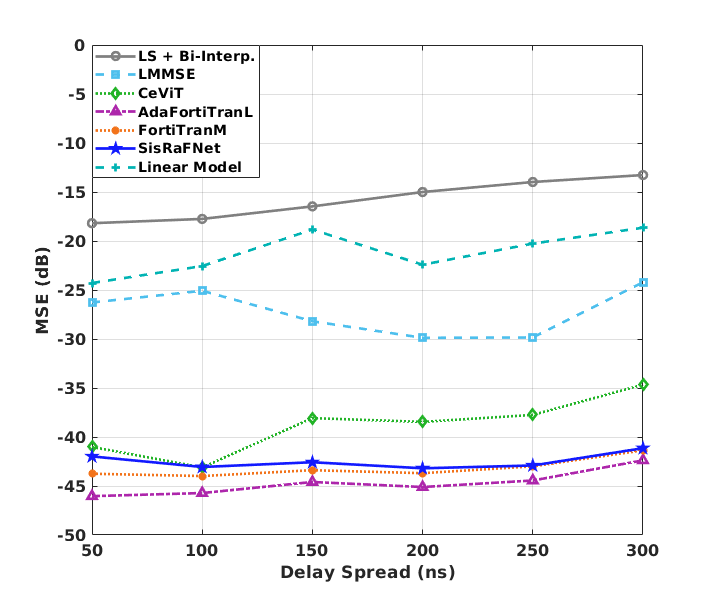}
        \caption{MSE vs. Delay Spread}
        \label{fig:ds}
    \end{subfigure}
    \hfill
    \begin{subfigure}{0.32\textwidth}
        \centering
        \includegraphics[width=\textwidth]{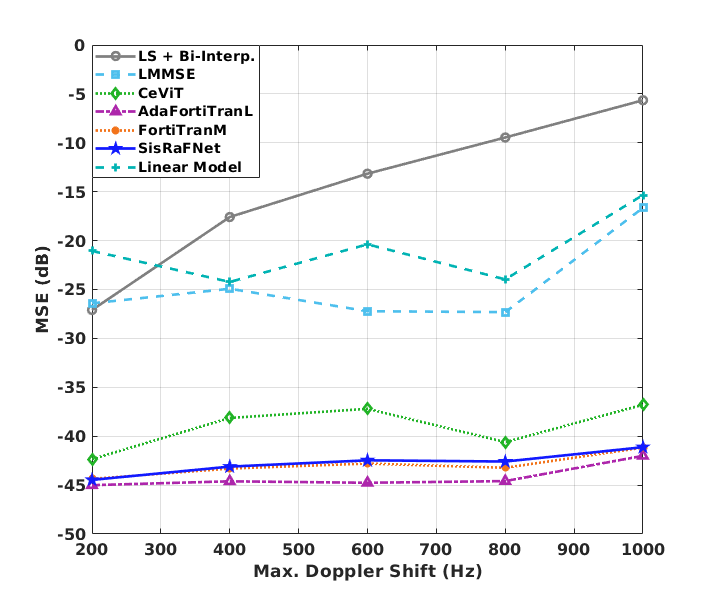}
        \caption{MSE vs. Maximum Doppler Shift}
        \label{fig:mds}
    \end{subfigure}
    \caption{Performance Analysis Across Diverse Channel Conditions}
    \label{fig:diverse}
\end{figure*}

\subsection{Datasets}

\subsubsection{Training and Validation Sets}
We generated 100,000 and 10,000 OFDM channel realizations for training and validation sets, respectively. The simulated channels follow the Tapped Delay Line-A (TDL-A) delay profile from the 3GPP specification \cite{3gpp_tdl}. For each channel realization, we select an SNR value from $\{0, 5, ..., 25\}$, a maximum Doppler shift value from $\{50, 100, ..., 1000\}$, and a delay spread value from $\{25, 50, ..., 300\}$, where SNR is in dB, Doppler shift is in Hz, and delay spread is in nanoseconds. We generate separate training and validation sets corresponding to pilot placements of shape ($40\times2$) for $N=3$, ($30\times2$) for $N=4$, ($24\times2$) for $N=5$, ($20\times2$) for $N=6$, and ($15\times2$) for $N=8$.

\subsubsection{Test Sets}
We generated several types of test sets, all based on TDL-A channels. The dynamic SNR test set comprises 2,000 test channels for each SNR value from $\{0, 5, ..., 25\}$, with delay spread set to 200 ns and maximum Doppler shift set to 500 Hz. The dynamic delay spread test set contains 2,000 test channels for each delay spread value from $\{50, 100, ..., 300\}$, with SNR and maximum Doppler shift fixed at 20 dB and 500 Hz, respectively. The dynamic Doppler shift dataset consists of 2,000 test channels for each maximum Doppler shift value from $\{200, 400, ..., 1000\}$.  For dynamic test sets, we used a pilot shape of ($40\times2$) corresponding to $N=3$.

The pilot density test set includes 2,000 channels for each pilot configuration, with SNR, maximum Doppler shift, and delay spread fixed at 5 dB, 500 Hz, and 200 ns, respectively. We consider pilot shapes of ($30\times2$), ($24\times2$), ($20\times2$), and ($15\times2$).

\subsection{Training}
AdaFortiTran and the linear model are trained with the Adam optimizer \cite{adam} with $\beta_1=0.9$ and $\beta_2=0.999$ for 1,000 epochs with early stopping to prevent overfitting. The batch size and initial learning rate are set to 512 and 0.001, respectively. We employ an exponential learning rate decay with a decay rate of 0.995 applied to every epoch. We use the above training sets to calculate channel statistics and noise variance for LMMSE. We implement and train Ce-ViT and SisRafNet as described in their respective papers \cite{tf3, bigru}. For training details not included in their papers, we follow the same procedure used for AdaFortiTran.

\subsection{Analysis}
\subsubsection{Robustness to Diverse Channel Scenarios}
We demonstrate the robustness of our proposed model by using the dynamic SNR, dynamic delay spread, and dynamic Doppler shift test sets. 

As shown in Fig. \ref{fig:snr}, AdaFortiTran outperforms its closest competitors with a 6 dB decrease in MSE at very low SNRs. The performance gap between Ce-ViT and AdaFortiTran remains constant around 6 dB under varying SNR. Similarly, our non-adaptive model FortiTran performs around 1 dB better than SisRafNet while having less than one-third of SisRafNet's size. Interestingly, FortiTran outperforms Ce-ViT at almost all SNR levels, achieving approximately 5 dB improvement at 25 dB SNR, despite Ce-ViT's knowledge of channel statistics. These results demonstrate the superiority of the proposed model in both adaptive and non-adaptive settings. Channel adaptivity guides our model to adjust its weights to challenging SNR conditions, as the performance gap between FortiTran and AdaFortiTran widens as SNR decreases. Also, it is interesting that LMMSE and AdaFortiTran behave very similarly in terms of their curve patterns. We attribute this to LMMSE taking the noise variance and channel correlations into account like AdaFortiTran does, showing that the learned behavior of AdaFortiTran in noise matches LMMSE's behavior.

In the dynamic delay spread test shown in Fig. \ref{fig:ds}, our proposed models continue to perform consistently better. Unlike the SNR-varying case, we do not observe typical behaviors among the adaptive and non-adaptive models. Another observation is that the performance gap between AdaFortiTran and FortiTran remains similar even for higher delay spread, suggesting that channel adaptivity does not help with dealing with delay spread as much as it does with SNR. Additionally, Ce-ViT does not seem to be as robust as other models despite its channel adaptivity, indicating that its architecture may not be fully utilizing this adaptive capability. We observe a similar pattern in Fig. \ref{fig:mds} where CE-ViT's performance oscillates more as the maximum Doppler shift value changes, showing that AdaFortiTran's adaptivity is more robust to changes in spectral and temporal characteristics. Furthermore, while the interpolated LS estimate shows an abrupt error increase as the Doppler shift increases, AdaFortiTran's error rate remains almost fixed, once again, demonstrating its robustness.

\subsubsection{Robustness to Pilot Density}
\begin{figure}[t]
    \centering
    \includegraphics[width=\columnwidth]{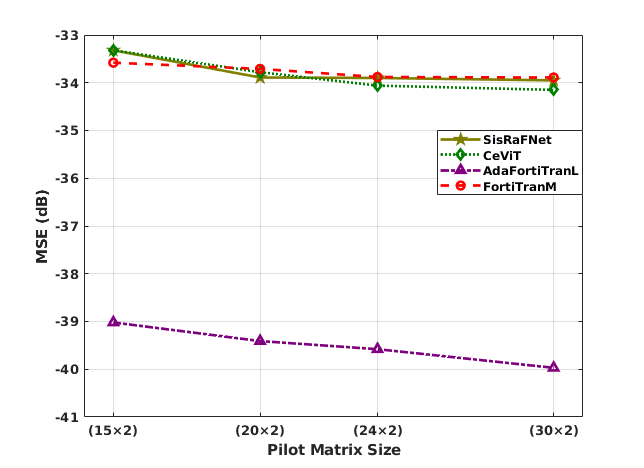}
    \caption{MSE vs Pilot Placement}
    \label{fig:pilot} 
\end{figure}

Fig. \ref{fig:pilot} displays the effect of pilot numbers on the channel estimation error. All models achieve lower error rates as the number of pilots increases. AdaFortiTran's superior performance persists despite changing pilot patterns, demonstrating that our model's superiority does not depend on pilot placement.

\section{Conclusion}
In this work, we proposed a compact transformer-based novel architecture that outperforms state-of-the-art deep learning channel estimation methods. Our method combined the strengths of several deep learning components and effectively utilized domain-specific knowledge. The extensive experimental results demonstrated that our approach maintains superior performance across varying SNR levels, delay spreads, and Doppler shifts, making it particularly suitable for practical wireless communications systems. The proposed architecture's robustness to different channel conditions, combined with its compact design, presents a promising solution for real-world deployments where computational resources may be limited. Our work demonstrates that carefully designed neural architectures incorporating domain knowledge can significantly advance the state of channel estimation, potentially leading to more efficient and reliable wireless communication systems.

\bibliographystyle{IEEEtran}  % Or any other style you prefer
\bibliography{references}  % Use the name of your .bib file without the .bib extension

\end{document}